\documentclass{article}

\usepackage{microtype}
\usepackage{graphicx}
\usepackage{subfigure}
\usepackage{booktabs} 
\usepackage{hyperref}

\usepackage[utf8]{inputenc} 
\usepackage[T1]{fontenc}    
\usepackage{hyperref}       
\usepackage{url}            
\usepackage{booktabs}       
\usepackage{amsfonts}       
\usepackage{nicefrac}       
\usepackage{microtype}      
\usepackage{amsmath,amsfonts,amsthm}
\usepackage{graphicx}

\DeclareMathOperator*{\diag}{diag}
\usepackage{tikz}

\usepackage{hyperref}

\usepackage[nonatbib,final]{neurips_2019}
\usepackage[numbers,square]{natbib}

\title{Deep Random Splines for Point Process Intensity Estimation of Neural Population Data}

\author{%
  Gabriel Loaiza-Ganem\\
  Department of Statistics\\
  Columbia University\\
  \texttt{gl2480@columbia.edu} \\
  \And
Sean M. Perkins \\
Department of Biomedical Engineering \\
Columbia University\\
  \texttt{sp3222@columbia.edu} \\
  \And
Karen E. Schroeder \\
Department of Neuroscience \\
Columbia University\\
  \texttt{ks3381@columbia.edu} \\
   \And
 Mark M. Churchland\\
  Department of Neuroscience\\
  Columbia University\\
  \texttt{mc3502@columbia.edu} \\ 
    \And
 John P. Cunningham\\
  Department of Statistics\\
  Columbia University\\
  \texttt{jpc2181@columbia.edu} \\ 
}

\begin{document}



\maketitle

\begin{abstract}
Gaussian processes are the leading class of distributions on random functions, but they suffer from well known issues including difficulty scaling and inflexibility with respect to certain shape constraints (such as nonnegativity). Here we propose Deep Random Splines, a flexible class of random functions obtained by transforming Gaussian noise through a deep neural network whose output are the parameters of a spline. Unlike Gaussian processes, Deep Random Splines allow us to readily enforce shape constraints while inheriting the richness and tractability of deep generative models. We also present an observational model for point process data which uses Deep Random Splines to model the intensity function of each point process and apply it to neural population data to obtain a low-dimensional representation of spiking activity. Inference is performed via a variational autoencoder that uses a novel recurrent encoder architecture that can handle multiple point processes as input. We use a newly collected dataset where a primate completes a pedaling task, and observe better dimensionality reduction with our model than with competing alternatives.\footnote{Our code is available at \url{https://github.com/cunningham-lab/drs}.}
\end{abstract}

\section{Introduction}

Gaussian Processes (GPs) are one of the main tools for modeling random functions \citep{rasmussen2004gaussian}. They allow control of the smoothness of the function by choosing an appropriate kernel but have the disadvantage that, except in special cases (for example \citet{gilboa2015scaling, flaxman2015fast}), inference in GP models scales poorly in both memory and runtime. Furthermore, GPs cannot easily handle shape constraints. It can often be of interest to model a function under some shape constraint, for example nonnegativity, monotonicity or convexity/concavity \citep{moller1998log, schmidt1988positivity, ramsay1988monotone, mammen1991estimating}. While some shape constraints can be enforced by transforming the GP or by enforcing them at a finite number of points, doing so cannot always be done and usually makes inference harder, see for example \citet{lin2014bayesian}.

Splines are another popular tool for modeling unknown functions \citep{wahba1990spline}. When there are no shape constraints, frequentist inference is straightforward and can be performed using linear regression, by writing the spline as a linear combination of basis functions. Under shape constraints, the basis function expansion usually no longer applies, since the space of shape constrained splines is not typically a vector space. However, the problem can usually still be written down as a tractable constrained optimization problem \citep{schmidt1988positivity}. Furthermore, when using splines to model a random function, a distribution must be placed on the spline's parameters, so the inference problem becomes Bayesian. \citet{dimatteo2001bayesian} proposed a method to perform Bayesian inference in a setting without shape constraints, but the method relies on the basis function expansion and cannot be used in a shape constrained setting. Furthermore, fairly simple distributions have to be placed on the spline parameters for their approximate posterior sampling algorithm to work adequately, which results in the splines having a restrictive and oversimplified distribution.

On the other hand, deep probabilistic models take advantage of the major progress in neural networks to fit rich, complex distributions to data in a tractable way \citep{rezende2014stochastic, mohamed2016learning, kingma2013auto, gao2016linear, johnson2016composing}. However, their goal is not usually to model random functions.

In this paper, we introduce Deep Random Splines (DRS), an alternative to GPs for modeling random functions. DRS are a deep probabilistic model in which standard Gaussian noise is transformed through a neural network to obtain the parameters of a spline, and the random function is then the corresponding spline. This combines the complexity of deep generative models and the ability to enforce shape constraints of splines.

We use DRS to model the nonnegative intensity functions of Poisson processes \citep{kingman1992poisson}. In order to ensure that the splines are nonnegative, we use a parameterization of nonnegative splines that can be written as an intersection of convex sets, and then use the method of alternating projections \citep{von1950geometry} to obtain a point in that intersection (and differentiate through that during learning). To perform scalable inference, we use a variational autoencoder \citep{kingma2013auto} with a novel encoder architecture that takes multiple, truly continuous point processes as input (not discretized in bins, as is common).

Our contributions are: $(i)$ Introducing DRS, $(ii)$ using the method of alternating projections to constrain splines, $(iii)$
proposing a variational autoencoder model whith a novel encoder architecture for point process data which uses DRS, and $(iv)$ showing that our model outperforms commonly used alternatives in both simulated and real data.

The rest of the paper is organized as follows: we first explain DRS, how to parameterize them and how constraints can be enforced in section \ref{DRS}. We then present our model and how to do inference in section \ref{obsmod}. We then compare our model against competing alternatives in simulated data and in two real spiking activity datasets, one of which we collected, in section \ref{expe}, and observe that our method outperforms the alternatives. Finally, we summarize our work in section \ref{conc}.

\section{Deep Random Splines}\label{DRS}

Throughout the paper we will consider functions on the interval $[T_1,T_2)$ and will select $I+1$ fixed knots $T_1=t_0<\dots<t_I=T_2$. We will refer to a function as a spline of degree $d$ and smoothness $s<d$ if the function is a $d$-degree polynomial in each interval $[t_{i-1}, t_i)$ for $i=1,\dots, I$, is continuous, and $s$ times differentiable. We will denote the set of splines of degree $d$ and smoothness $s$ by $\mathcal{G}_{d,s}=\{g_\psi:\psi\in\Psi_{d,s}\}$, where $\Psi_{d,s}$ is the set of parameters of each polynomial in each interval. That is, every $\psi \in \Psi_{d,s}$ contains the parameters of each of the $I$ polynomial pieces (it does not contain the locations of the knots as we take them to be fixed since we observed overfitting when not doing so). While the most natural ways to parameterize splines of degree $d$ are a linear combination of basis functions or with the $d+1$ polynomial coefficients of each interval, these parameterizations do not lend themselves to easily enforce constraints such as nonnegativity \citep{schmidt1988positivity}. We will thus use a different parameterization which we will explain in detail in the next section. We will denote by $\Psi \subseteq \Psi_{d,s}$ the subset of spline parameters that result in the splines having the shape constraint of interest, for example, nonnegativity.

DRS are a distribution over $\mathcal{G}_{d,s}$. To sample from a DRS, a standard Gaussian random variable $Z \in \mathbb{R}^m$ is transformed through a neural network parameterized by $\theta$, $f_\theta:\mathbb{R}^m \rightarrow \Psi$. The DRS is then given by $g_{f_\theta(Z)}$ and inference on $\theta$ can be performed through a variational autoencoder \citep{kingma2013auto}. Note that $f$ maps to $\Psi$, thus ensuring that the spline has the relevant shape constraint.

\subsection{Constraining Splines}\label{consplines}

We now explain how we can enforce piecewise polynomials to form a nonnegative spline. We add the nonnegativity constraint to the spline as we will use it for our model in section \ref{obsmod}, but constraints such as monotonicity and convexity/concavity can be enforced in an analogous way. In order to achieve this, we use a parameterization of nonnegative splines that might seem overly complicated at first. However, it has the critical advantage that it decomposes into the intersection of convex sets that are easily characterized in terms of the parameters, which is not the case for the naive parameterization which only includes the $d+1$ coefficients of every polynomial. We will see how to take advantage of this fact in the next section.

A beautiful but perhaps lesser known spline result (see \citet{lasserre2010moments}) gives that a polynomial $p(t)$ of degree $d$, where $d = 2k +1$ for some $k \in \mathbb{N}$, is nonnegative in the interval $[l,u)$ if and only if it can be written down as follows:
\begin{equation}\label{pospol}
p(t) = (u-t) [t]^\top Q_1[t] + (t-l) [t]^\top Q_2 [t]
\end{equation}
where $[t] = (1,t,t^2,\dots,t^k)^\top$ and $Q_1$ and $Q_2$ are $(k+1)\times (k+1)$ symmetric positive semidefinite matrices. It follows that a piecewise polynomial of degree $d$ with knots $t_0,\dots,t_I$ defined as $p^{(i)}(t)$ for $t \in [t_{i-1},t_i)$ for $i=1,\dots,I$ is nonnegative if and only if it can be written as:
\begin{equation}\label{posspline}
p^{(i)}(t) = (t_i-t) [t]^\top Q_1^{(i)}[t] + (t-t_{i-1}) [t]^\top Q_2^{(i)} [t]
\end{equation}
for $i=1,\dots,I$, where each $Q_1^{(i)}$ and $Q_2^{(i)}$ are $(k+1)\times (k+1)$ symmetric positive semidefinite matrices. We can thus parameterize every piecewise nonnegative polynomial on our $I$ intervals with $(Q_1^{(i)}, Q_2^{(i)})_{i=1}^I$. If no constraints are added on these parameters, the resulting piecewise polynomial might not be smooth, so certain constraints have to be enforced in order to guarantee that we are parameterizing a nonnegative spline and not just a nonnegative piecewise polynomial. To that end, we define $\mathcal{C}_1$ as the set of $(Q_1^{(i)}, Q_2^{(i)})_{i=1}^I$ such that:
\begin{equation}
p^{(i)}(t_{i}) = p^{(i+1)}(t_{i}) \text{ for } i=1,\dots, I-1
\end{equation}
that is, $\mathcal{C}_1$ is the set of parameters whose resulting piecewise polynomial as in equation \ref{posspline} is continuous. Analogously, let $\mathcal{C}_j$ for $j=2,3,\dots$ be the set of $(Q_1^{(i)}, Q_2^{(i)})_{i=1}^I$ such that:
\begin{equation}
\dfrac{\partial^{j-1}}{\partial t^{j-1}}p^{(i)}(t_{i}) = \dfrac{\partial^{j-1}}{\partial t^{j-1}}p^{(i+1)}(t_{i}) \text{ for } i=1,\dots, I-1
\end{equation}
so that $\mathcal{C}_j$ is the set of parameters whose corresponding piecewise polynomials have matching left and right $(j-1)$-th derivatives. Let $\mathcal{C}_0$ be the set of $(Q_1^{(i)}, Q_2^{(i)})_{i=1}^I$ which are symmetric positive semidefinite. We can then parameterize the set of nonnegative splines on $[T_1,T_2)$ by $\Psi= \cap_{j=0}^{s+1} \mathcal{C}_j$. Note that the case where $d$ is even can be treated analogously (see appendix 1).

\subsection{The Method of Alternating Projections}\label{altproj}

In order to use a DRS, $f_\theta$ has to map to $\Psi$, that is, we need to have a way for a neural network to map to the parameter set corresponding to nonnegative splines. We achieve this by taking $f_\theta (z) = h(\tilde{f}_\theta (z))$, where $\tilde{f}_\theta$ is an arbitrary neural network and $h$ is a surjective function onto $\Psi$. The most natural choice for $h$ is the projection onto $\Psi$. However, while computing the projection onto $\Psi$ (for $\Psi$ as in section \ref{consplines}) can be done by solving a convex optimization problem, it cannot be done analytically. This is an issue because when we train the model, we will need to differentiate $f_\theta$ with respect to $\theta$. Note that \citet{amos2017optnet} propose a method to have an optimization problem as a layer in a neural network. One might hope to use their method for our problem, but it cannot be applied due to the semidefinite constraint on our matrices. Concurrently to our work, \citet{agrawal2019differentiable} developed a method to differentiate through convex optimization problems that is compatible with our semidefinite constraints. We leave further exploration of their method within our framework for future work.

The method of alternating projections \citep{von1950geometry, bauschke1996projection} allows us to approximately compute such a function $h$ analytically. If $\mathcal{C}_0,\dots,\mathcal{C}_{s+1}$ are closed, convex sets in $\mathbb{R}^D$, then the sequence $\psi^{(k)} = P_{k\text{ mod }(s+2)}(\psi^{(k-1)})$ converges to a point in $\cap_{j=0}^{s+1} \mathcal{C}_j$ for any starting $\psi^{(0)}$, where $P_j$ is the projection onto $\mathcal{C}_j$ for $j=0,\dots,s+1$. The method of alternating projections then consists on iteratively projecting onto each set in a cyclic fashion. We call computing $\psi^{(k)}$ from $\psi^{(k-1)}$ the $k$-th iteration of the method of alternating projections. This method can be useful to obtain a point in the intersection if each $P_j$ can be easily computed.

In our case, projecting onto $\mathcal{C}_0$ can be done by doing eigenvalue decompositions of $Q_1^{(i)}$ and $Q_2^{(i)}$ and zeroing out negative elements in the diagonal matrices containing the eigenvalues. While this projection might seem computationally expensive, the matrices are small and this can be done efficiently. For example, for cubic splines ($d=3$), there are $2I$ matrices each one of size $2\times2$. Projecting onto $\mathcal{C}_j$ for $j=1,\dots s+1$ can be done analytically as it can be formulated as a quadratic optimization problem with linear constraints. Furthermore, because of the local nature of the constraints where every interval is only constrained by its neighboring intervals, this quadratic optimization problem can be reduced to solving a tridiagonal system of linear equations of size $I-1$ which can be solved efficiently in $O(I)$ time with simplified Gaussian elimination. We prove this fact, using the KKT conditions, in appendix 2.

By letting $h$ be the first $M$ iterations of the method of alternating projections, we can ensure that $f_\theta$ maps (approximately) to $\Psi$, while still being able to compute $\nabla_\theta f_\theta(z)$. Note that we could find such an $h$ function using Dykstra's algorithm (not to be confused with Dijkstra's shortest path algorithm), which is a modification of the method of alternating projections that converges to the projection of $\psi^{(0)}$ onto $\cap_{j=0}^{s+1} \mathcal{C}_j$ \citep{dykstra1983algorithm, boyle1986method, tibshirani2017dykstra}), but we found that the method of alternating projections was faster to differentiate when using reverse mode automatic differentiation packages \citep{abadi2016tensorflow}.

Another way of finding such an $h$ would be unrolling  any iterative optimization method that solves the projection onto $\Psi$, such as gradient-based methods or Newton methods. We found the alternating projections method more convenient as it does not involve additional hyperparameters such as learning rate that drastically affect performance. Furthermore, the method of alternating projections is known to have a linear convergence rate (as fast as gradient-based methods) that is independent of the starting point \citep{bauschke1996projection}. This last observation is important, as the starting point in our case is determined by the output of $\tilde{f}_\theta$, so that the convergence rate being independent of the starting point ensures that $\tilde{f}_\theta$ cannot learn to ignore $h$, which is not the case for gradient-based and Newton methods (for a fixed number of iterations and learning rate, there might exist an initial point that is too far away to actually reach the projection). Finally, note that if we wanted to enforce, for example, that the spline be monotonic, we could parameterize its derivative and force it to be nonnegative or nonpositive. Convexity or concavity can be enforced analogously.

\section{Deep Random Splines as Intensity Functions of Point Processes}\label{obsmod}

Since we will use DRS as intensity functions for Poisson processes, we begin this section with a brief review of these processes.

\subsection{Poisson Processes}
An inhomogeneous Poisson process in a set $\mathcal{S}$ is a random subset of $\mathcal{S}$. The process can (for our purposes) be parameterized by an intensity function $g: \mathcal{S} \rightarrow \mathbb{R}_{+}$ and in our case,  $\mathcal{S}=[T_1,T_2)$. We write $S \sim \mathcal{PP}_{\mathcal{S}}(g)$ to denote that the random set $S$, whose elements we call events, follows a Poisson process on $\mathcal{S}$ with intensity $g$. If $S =\{x_k\}_{k=1}^K \sim \mathcal{PP}_{\mathcal{S}}(g)$, then $|S \cap A|$, the number of events in any $A \subseteq \mathcal{S}$, follows a Poisson distribution with parameter $\int_A g(t)dt$ and the log likelihood of $S$ is given by:
\begin{equation}\label{pp_log_lik}
\log p(\{x_k\}_{k=1}^{K} | g) = \displaystyle \sum_{k=1}^K \log g(x_k) - \int_{\mathcal{S}} g(t)dt
\end{equation}
Splines have the very important property that they can be analytically integrated (as the integral of polynomials can be computed in closed form), which allows to exactly evaluate the log likelihood in equation \ref{pp_log_lik} when $g$ is a spline. As a consequence, fitting a DRS to observed events is more tractable than fitting models that use GPs to represent $g$, such as log-Gaussian Cox processes \citep{moller1998log}. Inference in the latter type of models is very challenging, despite some efforts by \citet{cunningham2008fast, adams2009tractable, lloyd2015variational}. Splines also vary smoothly, which incorporates the reasonable assumption that the expected number of events changes smoothly over time. These properties were our main motivations for choosing splines to model intensity functions.

\subsection{Our Model}\label{ppvae}

Suppose we observe $N$ simultaneous point processes in $[T_1,T_2)$ a total of $R$ repetitions (we will call each one of these repetitions/samples a trial). Let $X_{r,n}$ denote the $n$-th point process of the $r$-th trial. Looking ahead to an application we study in the results, data of this type is a standard setup for microelectrode array data, where $N$ neurons are measured from time $T_1$ to time $T_2$ for $R$ repetitions, and each event in the point processes corresponds to a spike (the time at which the neurons ``fired''). Each $X_{r,n}$ is also called a spike train. The model we propose, which we call DRS-VAE, is as follows:
\begin{figure}[h]
\begin{minipage}{0.5\textwidth}
\begin{equation}\label{model}
\begin{cases}
Z_r \sim \mathcal{N}(0, I_m) \text{ for }r=1,\dots,R\\
\psi_{r,n} = f_{\theta}^{(n)}(Z_r) \text{ for }n=1,\dots,N\\
X_{r,n}|\psi_{r,n} \sim \mathcal{PP}_{[T_1,T_2)}(g_{\psi_{r,n}})
\end{cases}
\end{equation}
\end{minipage}
\begin{minipage}{0.5\textwidth}
\centering
\includegraphics[scale=0.7, trim={100 530 250 125},clip]{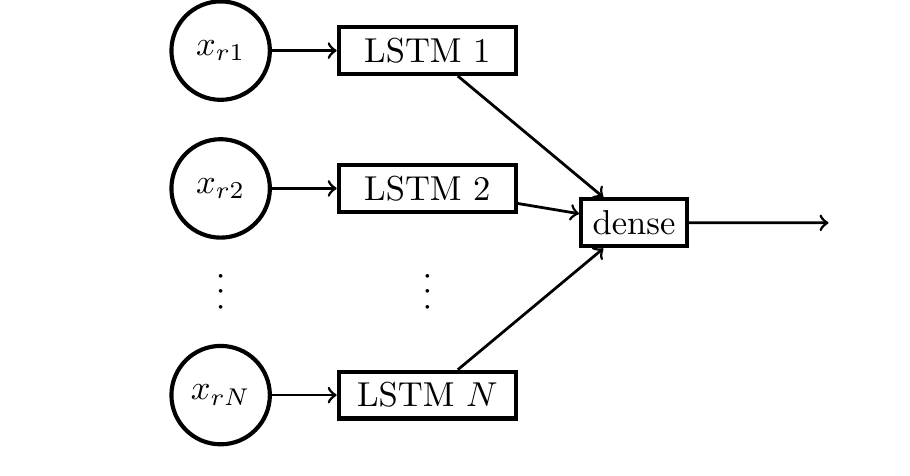}
  \caption{Encoder architecture.}
  \label{fig:LSTMarch}
\end{minipage}
\hspace{-35pt}
\end{figure}

where each $f_\theta^{(n)}:\mathbb{R}^m\rightarrow \Psi$ is obtained as described in section \ref{altproj}. The hidden state $Z_r$ for the $r$-th trial $\bold{X}_r:=(X_{r,1},\dots, X_{r,N})$ can be thought as a low-dimensional representation of $\bold{X}_r$. Note that while the intensity function of every point process and every trial is a DRS, the latent state $Z_r$ of each trial is shared among the $N$ point processes. Note also that the data we are modeling can be thought of as $R$ marked point processes \citep{kingman1992poisson}, where the mark of the event $x_{r,n,k}$ (the $k$-th event of the $n$-th point process of the $r$-th trial) is $n$. In this setting, $g_{\psi_{r,n}}$ corresponds to the conditional (on $Z_r$ and on the mark being $n$) intensity of the process for the $r$-th trial.

Once again, one might think that our parameterization of nonnegative splines is unnecessarily complicated and that having $f_{\theta}^{(n)}$ in equation \ref{model} be a simpler parameterization of an arbitrary spline (e.g. basis coefficients) and using $\tau(g_{\psi_{r,n}})$ instead of $g_{\psi_{r,n}}$, where $\tau$ is a nonnegative function, might be a better solution to enforcing nonnegativity constraints. The function $\tau$ would have to be chosen in such a way that the integral of equation \ref{pp_log_lik} can still be computed analytically, making $\tau(t) = t^2$ a natural choice. While this procedure would avoid having to use the method of alternating projections, we found that squared splines perform very poorly as they oscillate too much. Alternatively, we also tried using a B-spline basis with nonnegative coefficients, resulting in nonnegative splines. While the approximation error between a nonnegative smooth function and its B-spline approximation with nonnegative coefficients can be bounded \citep{shen2016adaptive}, note that not every nonnegative spline can be written down as a linear combination of B-splines with nonnegative coefficients. In practice we found the bound to be too loose and also obtained better performance through the method of alternating projections.

\subsection{Inference}

Autoencoding variational Bayes \citep{kingma2013auto} is a technique to perform inference in the following type of model:
\begin{equation}
\begin{cases}
Z_r \sim  \mathcal{N}(0, I_m) \text{ for }r=1,\dots,R\\
X_r | Z_r \sim p_\theta(x|z_r)
\end{cases}
\end{equation}
where $X_r$ are the observables and $Z_r \in \mathbb{R}^m$ the corresponding latents. Since maximum likelihood is not usually tractable, the posterior $p(\bold{z}|\bold{x})$ is approximated with $q_\phi(\bold{z}|\bold{x})$, which is given by:
\begin{equation}\label{mean_field_vi}
q_\phi(\bold{z}|\bold{x}) = \displaystyle \prod_{r=1}^R q_\phi(z_r|x_r), \text{with }q_\phi(z_r|x_r) = \mathcal{N}\Big(\mu_\phi(x_r), \diag\big(\sigma^2_\phi(x_r)\big)\Big)
\end{equation}
where the encoder $(\mu_\phi, \sigma_\phi)$ is a neural network parameterized by $\phi$. The ELBO $\mathcal{L}$, a lower bound of the log likelihood, is then maximized over both the generative parameters $\theta$ and the variational parameters $\phi$:
\begin{equation}\label{ELBO1}
\mathcal{L}(\theta, \phi)  = \displaystyle \sum_{r=1}^R -KL(q_\phi(z_r|x_r)|| p(z_r)) + E_{q_\phi(z_r|x_r)}[\log p_\theta(x_r | z_r)]
\end{equation}
Maximizing the ELBO with stochastic gradient methods is enabled by the use of the reparameterization trick. In order to perform inference in our model, we use autoencoding variational Bayes. Because of the point process nature of the data, $\mu_\phi$ and $\sigma_\phi$ require a recurrent architecture, since their input $\bold{x}_r=(x_{r,1}, x_{r,2},\dots, x_{r,N})$ consists of $N$ point processes. This input does not consist of a single sequence, but $N$ sequences of different lengths (numbers of events), which requires a specialized architecture. We use $N$ separate LSTMs \citep{hochreiter1997long}, one per point process. Each LSTM takes as input the events of the corresponding point process. The final states of each LSTM are then concatenated and transformed through a dense layer (followed by an exponential activation in the case of $\sigma_\phi$ to ensure positivity) in order to map to the hidden space $\mathbb{R}^m$. We also tried bidirectional LSTMs \citep{graves2005framewise} but found regular LSTMs to be faster while having similar performance. The architecture is depicted in figure \ref{fig:LSTMarch}. The ELBO for our model is then given by:
\small
\begin{equation}
\mathcal{L}(\theta, \phi) = \displaystyle \sum_{r=1}^R -KL(q_\phi(z_r|x_r)|| p(z_r)) + E_{q_\phi(z_r|x_r)}\Big[\sum_{n=1}^N \sum_{k=1}^{K_{r,n}} \log g_{\psi_{r,n}}(x_{r,n,k}) - \int_{\mathcal{S}} g_{\psi_{r,n}}(t)dt\Big]
\end{equation}
\normalsize
where $K_{r,n}$ is the number of events in the $n$-th point process of the $r$-th trial. \citet{gao2016linear} have a similar model, where a hidden Markov model is transformed through a neural network to obtain event counts on time bins. The hidden state for a trial in their model is then an entire hidden Markov chain, which will have significantly higher dimension than our hidden state. Also, their model can be recovered from ours if we change the standard Gaussian distribution of $Z_r$ in equation \ref{model} to reflect their Markovian structure and choose $\mathcal{G}$ to be piecewise constant, nonnegative functions. We also emphasize the fact that our model is very easy to extend: for example, it would be straightforward to extend it to multi-dimensional point processes (not neural data any more) by changing $\mathcal{G}$ and its parameterization. It is also straightforward to use a more complicated point process than the Poisson one by allowing the intensity to depend on previous event history. Furthermore, DRS can be used in settings that require random functions, even if no point process is involved.

One of the advantages of our method is that it scales well (not cubically, like most GP methods) with respect to most of its parameters like number of trials, number of knots, number of iterations of the alternating projections algorithm, hidden dimension and number of neurons. The only parameter with which our method does not scale as well is the number of spikes since the LSTM-based encoder has to process every spike individually (not spike counts over time bins). However, this issue can be addressed by using a non-amortized inference approach (i.e. not having an encoder and having separate variational parameters for each trial). We found that the amortized approach using our proposed encoder was better for the datasets we analyzed, but even larger datasets might benefit from the non-amortized approach.


%
%

\section{Experiments}\label{expe}

\subsection{Simulated Data}

We simulated data with the following procedure: First, we set $2$ different types of trials. For each type of trial, we sampled one true intensity function on $[0, 10)$ for each of the $N=2$ point processes by sampling from a GP and exponentiating the result. We then sampled 600 times from each type of trial, resulting in 1200 trials. We randomly selected 1000 trials for training and set aside the rest for testing. We then fit the model described in section \ref{ppvae} and compare against other methods that perform intensity estimation while recovering a low-dimensional representation of trial: the PP-GPFA model \citep{duncker2018temporal}, the PfLDS model \citep{gao2016linear} and the GPFA model \citep{yu2009gaussian}. The two latter models discretize time into $B$ time bins and have a latent variable per time bin and per trial (as opposed to our model which is only per trial), while the former recovers continuous latent trajectories. They do this as a way of enforcing temporal smoothness by placing an appropriate prior over their latent trajectories, which we do not have to do as we implicitly enforce temporal smoothness by using splines to model intensity functions. Note that \citet{du2012learning}, \citet{yang2017decoupling}, \citet{mei2017neural} and \citet{du2016recurrent} all propose related methods in which the intensity of point processes is estimated. However, we do not compare against these as the two former ones model dynamic networks, making a direct comparison difficult, and the two latter do not use latent variables, which is one of the main advantages and goals of our method as a way to perform dimensionality reduction for neural population data.

We used a uniform grid with 11 knots (resulting in $I=10$ intervals), $d=3$ and $s=2$.  Since a twice-differentiable cubic spline on $I$ intervals has $I+3$ degrees of freedom, when discretizing time for PfLDS and GPFA we use $B=I+3=13$ time bins. This way the distribution recovered by PfLDS also has $B=13$ degrees of freedom, while the distribution recovered by GPFA has even more. We set the latent dimension $m$ in our model to $2$ and we also set the latent dimension per time bin in PfLDS and GPFA to $2$, meaning that the overall latent dimension for an entire trial was $2B = 26$. These two choices make the comparison conservative as they allow more flexibility for the two competing methods than for ours. For PP-GPFA we set the continuous latent trajectory to have dimension $2$. Our architecture and hyperparameter choices are included in appendix 3.

The top left panel of figure \ref{fig:comp} shows the posterior means of the hidden variables in our model for each of the $200$ test trials. Each posterior mean is colored according to its type of trial. We can see that different types of trials form separate clusters, meaning that our model successfully obtains low-dimensional representations of the trials. Note that the model is trained without having access to the type of each trial; colors are assigned in the figure post hoc. The top right panel shows the events (in black) for a particular point process on a particular trial, along with the true intensity (in green) that generated the events and posterior samples from our model (in purple), PP-GPFA (in orange), PfLDS (in blue), and GPFA (in red) of the corresponding intensities. Note that since PfLDS and GPFA parameterize the number of counts on each time bin, they do not have a corresponding intensity. We plot instead a piecewise constant intensity on each time bin in such a way that the expected number of events in each time bin is equal to the integral of the intensity. We can see that our method recovers a smooth function that is closer to the truth than the ones recovered with competing methods. The bottom left panel of figure \ref{fig:comp} further illustrates this point with a QQ-plot (where time is rescaled as in \cite{brown2002time}), and we can see once again that our method recovers intensities that are closer to the truth.

\begin{figure}
\begin{tabular}{cc}
\centering
\includegraphics[scale=0.52]{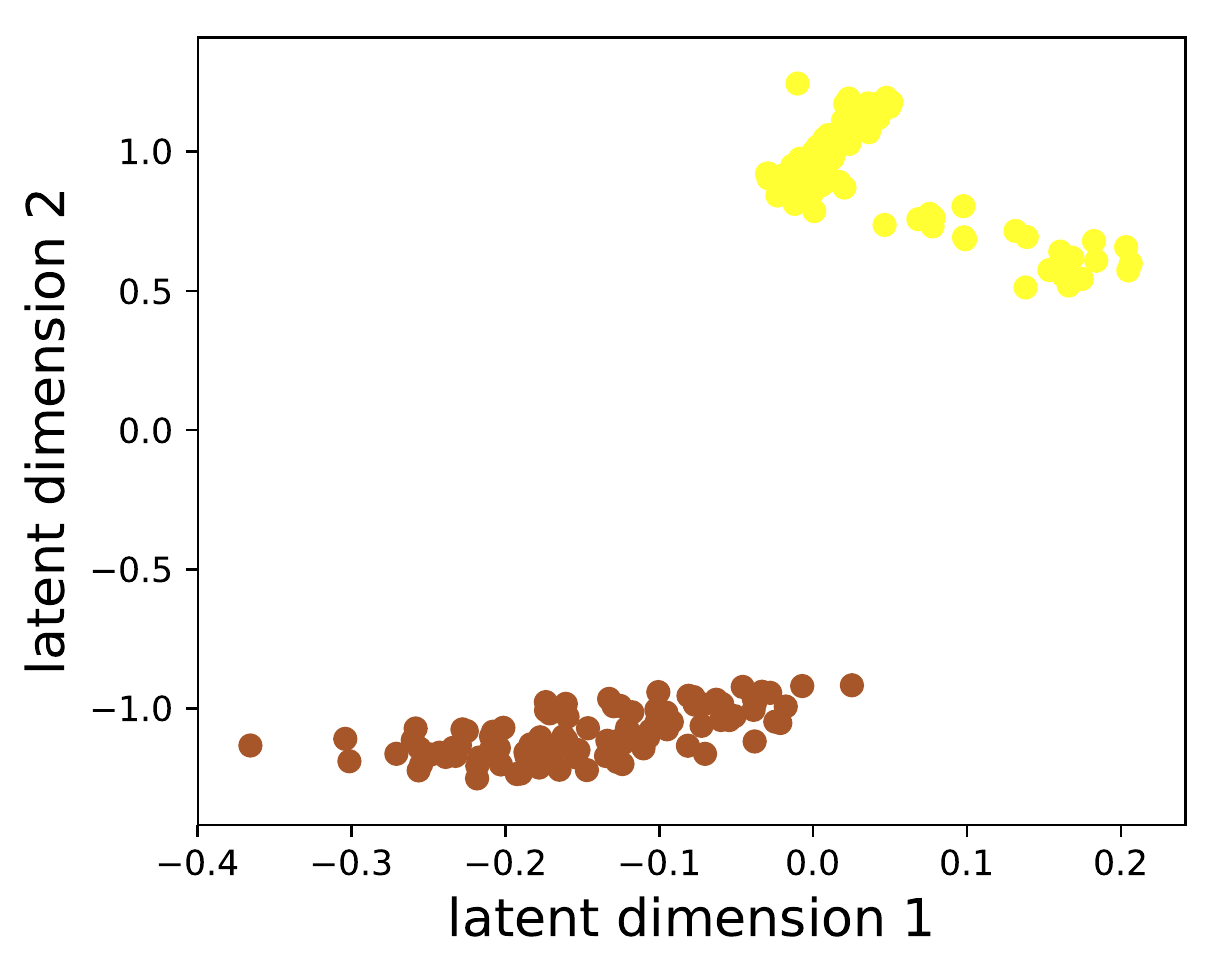} &
\includegraphics[scale=0.52]{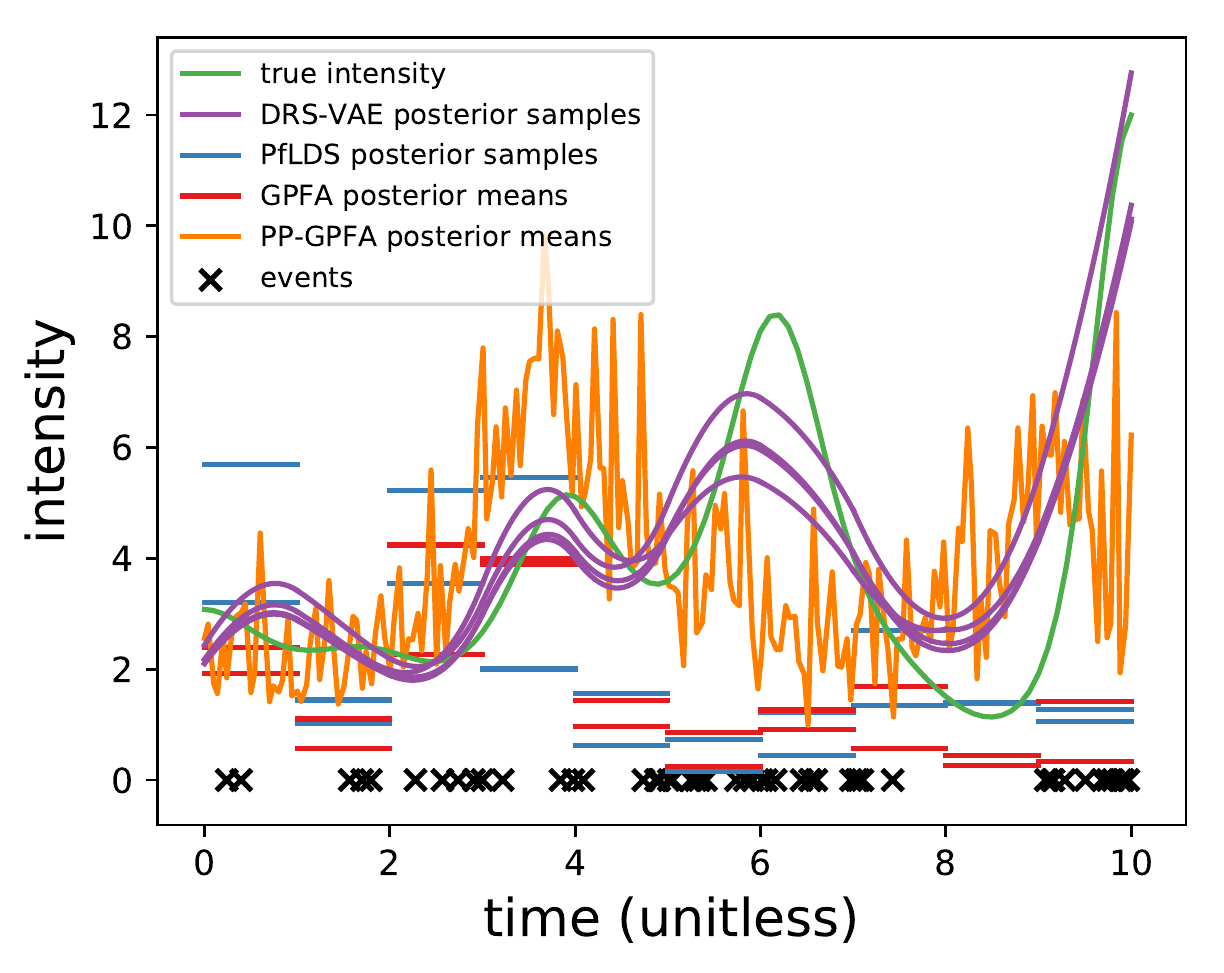}\\
\includegraphics[scale=0.52]{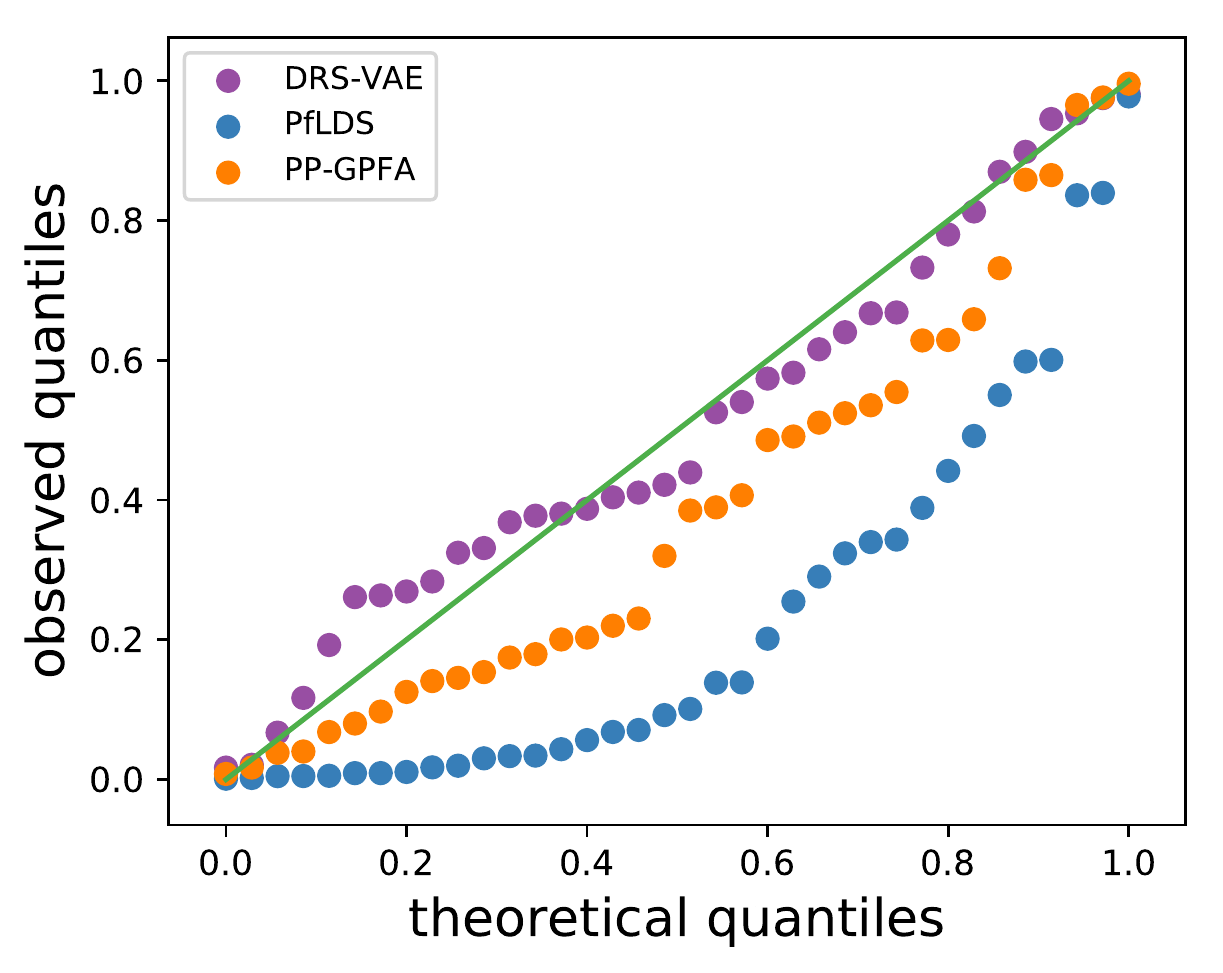}&
\includegraphics[scale=0.52]{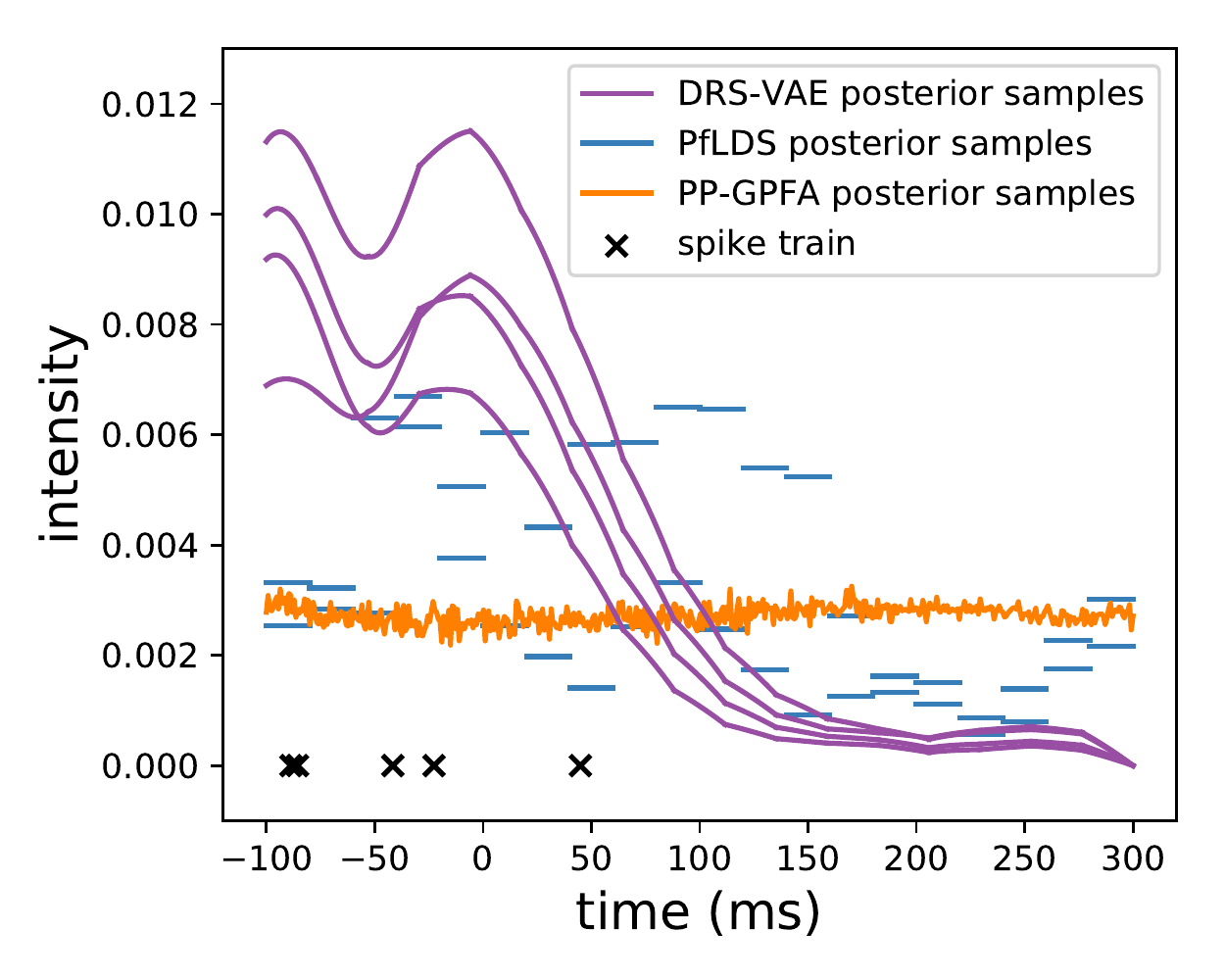}
\end{tabular}
\caption{Posterior means of the hidden variables of DRS-VAE by type of trial on simulated data (top left panel), QQ-plot of time-rescaled intensities on simulated data (bottom left panel), comparison of posterior intensities of our method (DRS-VAE) against competing alternatives on simulated data (top right panel) and reaching data (bottom right panel).}
\label{fig:comp}
\end{figure}


\begin{table}[b]
\caption{Quantitative comparison of our method (DRS-VAE) against competing alternatives.} \label{table:comparison}
\tiny
\begin{center}
\begin{tabular}{lccccccccc}
& \multicolumn{3}{c}{\footnotesize \textbf{SIMULATED DATA}} & \multicolumn{3}{c}{\footnotesize \textbf{REACHING DATA}} & \multicolumn{3}{c}{\footnotesize \textbf{CYCLING DATA}}\\
\cmidrule(r){2-4}\cmidrule(r){5-7}\cmidrule(r){8-10}
{\footnotesize \textbf{METHOD}}  &\textbf{ELBO}  &$\bold{L^2}$ & $\bold{p\text{\textbf{-VALUE}}}$ & \textbf{ELBO} & \textbf{15-NN} & \textbf{SSG/SST} &\textbf{ELBO}  & \textbf{15-NN} & \textbf{SSG/SST} \\
\hline \\
{\footnotesize DRS-VAE} & $\bold{57.1}$ & $\bold{4.43 \pm 3.55}$ & $-$ & $\bold{-500.8}$ & $\bold{23.7\%}$ & $\bold{73.9\%}$ & $6372$ & $\bold{55.9\%}$ & $\bold{70.0\%}$\\
{\footnotesize PfLDS} & $52.3$ & $11.9 \pm 6.18$ & $<10^{-73}$ & $-505.7$ & $3.1\%$ & $6.2\%$ & $\bold{6532}$ & $11.7\%$ & $3.2\%$\\
{\footnotesize GPFA} & $-$ & $12.9 \pm 7.19$ & $<10^{-72}$ & $-$ & $-$ & $-$ & $-$ & $-$ & $-$\\
{\footnotesize PP-GPFA} & $29.0$ & $15.4 \pm 9.64$ & $<10^{-71}$ & $-523.2$ & $14.1\%$ & $30.5\%$ & $6079$ & $51.1\%$ & $14.6\%$
\end{tabular}
\end{center}
\label{table:comp}
\end{table}

Table \ref{table:comp} shows performance from our model compared against PP-GPFA, PfLDS and GPFA. The second column shows the per-trial ELBO on test data, and we can see that our model has a larger ELBO than the alternatives. While having a better ELBO does not imply that our log likelihood is better, it does suggest that it is. Since both PfLDS and GPFA put a distribution on event counts on time bins instead of a distribution on event times as our models does, the log likelihoods are not directly comparable. However, in the case of PfLDS, we can easily convert from the Poisson likelihood on time bins to the piecewise constant intensity Poisson process likelihood, so that the numbers become comparable. In order to get a quantitative comparison between our model and GPFA, we take advantage of the fact that we know the true intensity that generated the data and compare average $L^2$ distance, across point processes and trials, between posterior intensity samples and actual intensity function. Once again, we can see that our method outperforms the alternatives. Table \ref{table:comp} also includes the standard deviation of these $L^2$ distances. Since the standard deviations are somewhat large in comparison to the means, for each of the two competing alternatives, we carry out a two sample t-test comparing the $L^2$ distance means obtained with our method against the alternative. The $p$-values indicate that our method recovers intensity functions that are closer to the truth in a statistically significant way.

\subsection{Real Data}

\subsubsection{Reaching Data}

We also fit our model to the dataset collected by \citet{churchland2012neural}. The dataset, after preprocessing (see appendix 4 for details), consists of measurements of $20$ neurons for $3590$ trials on the interval $[-100,300)$ (in $ms$) of a primate. In each trial, the primate reaches with its arm to a specific location, which changes from trial to trial (we can think of the $40$ locations as types of trials), where time $0$ corresponds to the beginning of the movement. We randomly split the data into a training set with $3000$ trials and a test set with the rest of the trials.

We used twice-differentiable cubic splines and $18$ uniformly spaced knots (that is, $17$ intervals). For the comparison against PfLDS, we split time into $20$ bins, resulting in time bins of $20ms$ (which is a standard length), once again making sure that the degrees of freedom are comparable. This makes once more for a conservative comparison as we fix the number of knots in our model so that the number of degrees of freedom match against the already tuned comparison instead of tuning the number of knots directly. Further architectural details are included in appendix 3. Since we do not have access to the ground truth, we do not compare against GPFA as the $L^2$ metric computed in the previous section cannot be used here. Again, we used a hidden dimension $m=2$ for our model, resulting in hidden trajectories of dimension $40$ for PfLDS, and continuous trajectories of dimension $2$ for PP-GPFA. We experimented with larger values of $m$ but did not observe significant improvements in either model.

The bottom right panel of figure \ref{fig:comp} shows the spike train (black) for a particular neuron on a particular trial, along with posterior samples from our model (in purple), PP-GPFA (in orange) and PfLDS (in blue) of the corresponding intensities. We can see that the posterior samples from our method look more plausible and smoother than the other ones.

Table \ref{table:comp} also shows the per-trial ELBO on test data for our model and for the competing alternatives. Again, our model has a larger ELBO, even when PfLDS has access to $20$ times more hidden dimensions: our method is more successful at producing low-dimensional representations of trials than PfLDS. The table also shows the percentage of correctly predicted test trial types when using $15$-nearest neighbors on the posterior means of train data (the entire trajectories are used for PfLDS and $20$ uniformly spaced points along each dimension of the continuous trajectories of PP-GPFA, resulting in $40$ dimensional latent representations). While $23.7\%$ might seem small, it should be noted that it is significantly better than random guessing (which would have $2.5\%$ accuracy) and that the model was not trained to minimize this objective. Regardless, we can see that our method outperforms both PP-GPFA and PfLDS in this metric, even when using a much lower-dimensional representation of each trial. The table also includes the percentage of explained variation when doing ANOVA on the test posterior means (denoted SSG/SST), using trial type as groups. Once again, we can see that our model recovers a more meaningful representation of the trials.

\subsubsection{Cycling Data}

We also fit our model to our newly collected dataset. After preprocessing (see supplementary material), it consists of $1300$ and $188$ train and test trials, respectively. During each trial, $20$ neurons were recorded as the primate turns a hand-held pedal to navigate through a virtual environment. There are 8 trial types, based on whether the primate is pedaling forward or backward and over what distance.

We use the same hyperparameter settings as for the reaching data, except we use $26$ uniformly spaced knots ($25$ intervals) and $28$ bins for PfLDS, as well as a hidden dimension $m=10$, resulting in hidden trajectories of dimension $280$ for PfLDS (analogously, we  set PP-GPFA to have $10$ dimensional continuous trajectories, and take $28$ uniformly spaced points along each dimension to obtain $280$ dimensional latent representations). Results are also summarized in table \ref{table:comp}. We can see that while our ELBO is higher than for PP-GPFA, it is actually lower than for PfLDS, which we believe is caused by an artifact of preprocessing the data rather than any essential performance loss.

While the ELBO was better for PfLDS, the quality of our latent representations is significantly better, as shown by the accuracy of $15$-nearest neighbors to predict test trial types (random guessing would have $12.5\%$ accuracy) and the ANOVA percentage of explained variation of the test posterior means, which are also better than for PP-GPFA. This is particularly impressive as our latent representations have $28$ times fewer dimensions.  We did experiment with different hyperparameter settings, and found that the ELBO of PfLDS increased slightly when using more time bins (at the cost of even higher-dimensional latent representations), whereas our ELBO remained the same when increasing the number of intervals. However, even in this setting the accuracy of $15$-nearest neighbors and the percentage of explained variation did not improve for PfLDS.

\section{Conclusions}\label{conc}

In this paper we introduced Deep Random Splines, an alternative to Gaussian processes to model random functions. Owing to our key modeling choices and use of results from the spline and optimization literatures, fitting DRS is tractable and allows one to enforce shape constraints on the random functions. While we only enforced nonnegativity and smoothness in this paper, it is straightforward to enforce constraints such as monotonicity (or convexity/concavity). We also proposed a variational autoencoder that takes advantage of DRS to accurately model and produce meaningful low-dimensional representations of neural activity.

Future work includes using DRS-VAE for multi-dimensional point processes, for example spatial point processes. While splines would become harder to use in such a setting, they could be replaced by any family of easily-integrable nonnegative functions, such as, for example, conic combinations of Gaussian kernels. Another line of future work involves using a more complicated point process than the Poisson, for example a Hawkes process, by allowing the parameters of the spline in a certain interval to depend on the previous spiking history of previous intervals. Finally, DRS can be applied in more general settings than the one explored in this paper since they can be used in any setting where a random function is involved, having many potential applications beyond what we analyzed here.

\subsubsection*{Acknowledgments}

We thank the Simons Foundation, Sloan Foundation, McKnight Endowment Fund, NIH NINDS 5R01NS100066, NSF 1707398, and the Gatsby Charitable Foundation for support.

\nocite{yuan2006model}
\bibliographystyle{abbrvnat}
\bibliography{bibliography}{}


\end{document}